\documentclass[10pt,twocolumn,letterpaper]{article}

\usepackage{cvpr}              

\usepackage{graphicx}
\usepackage{amsmath}
\usepackage{amssymb}
\usepackage{booktabs}

\usepackage[pagebackref,breaklinks,colorlinks]{hyperref}

\makeatletter
\renewcommand{\maketag@@@}[1]{\hbox{\m@th\normalsize\normalfont#1}}%
\makeatother

\usepackage[noend]{algpseudocode}
\usepackage{algorithmicx,algorithm}

\usepackage[capitalize]{cleveref}
\crefname{section}{Sec.}{Secs.}
\Crefname{section}{Section}{Sections}
\Crefname{table}{Table}{Tables}
\crefname{table}{Tab.}{Tabs.}

\def\cvprPaperID{11} 
\def\confName{CVPR}
\def\confYear{2023}

\begin{document}

\title{Gradient Attention Balance Network: Mitigating Face Recognition Racial Bias via Gradient Attention}

\author{
Linzhi Huang\textsuperscript{1} \\
\and Mei Wang\textsuperscript{1} \\
\and Jiahao	Liang\textsuperscript{1} \\
\and Weihong Deng\textsuperscript{1} \\
\and Hongzhi Shi\textsuperscript{1,2} \\
\and Dongchao Wen\textsuperscript{1,2\thanks{Corresponding author.}}  \\
\and Yingjie Zhang\textsuperscript{1,2} \\
\and Jian Zhao\textsuperscript{1,2} \\
\and {\textsuperscript{1}Inspur Electronic Information Industry Co., Ltd} 
\and {\textsuperscript{2}Shandong Massive Information Technology Research Institute} \\
 {\tt\small\{huanglinzhi\_1, wangmei\_1, goodlplus, dengweihong2020\}@foxmail.com} \\
 {\tt\small\{shihzh, wendongchao, zhangyj-s, zhao\_jian\}@inspur.com}
}

\maketitle


\begin{abstract}
Although face recognition has made impressive progress in recent years, we ignore the racial bias of the recognition system when we pursue a high level of accuracy. 
Previous work found that for different races, face recognition networks focus on different facial regions, and the sensitive regions of darker-skinned people are much smaller. 
Based on this discovery, we propose a new de-bias method based on gradient attention, called Gradient Attention Balance Network (GABN). 
Specifically, we use the gradient attention map (GAM) of the face recognition network to track the sensitive facial regions and make the GAMs of different races tend to be consistent through adversarial learning. 
This method mitigates the bias by making the network focus on similar facial regions. 
In addition, we also use masks to erase the Top-N sensitive facial regions, forcing the network to allocate its attention to a larger facial region. 
This method expands the sensitive region of darker-skinned people and further reduces the gap between GAM of darker-skinned people and GAM of Caucasians. 
Extensive experiments show that GABN successfully mitigates racial bias in face recognition and learns more balanced performance for people of different races.
\end{abstract}

\begin{figure}[t]
\centering
\includegraphics[width=0.95\linewidth]{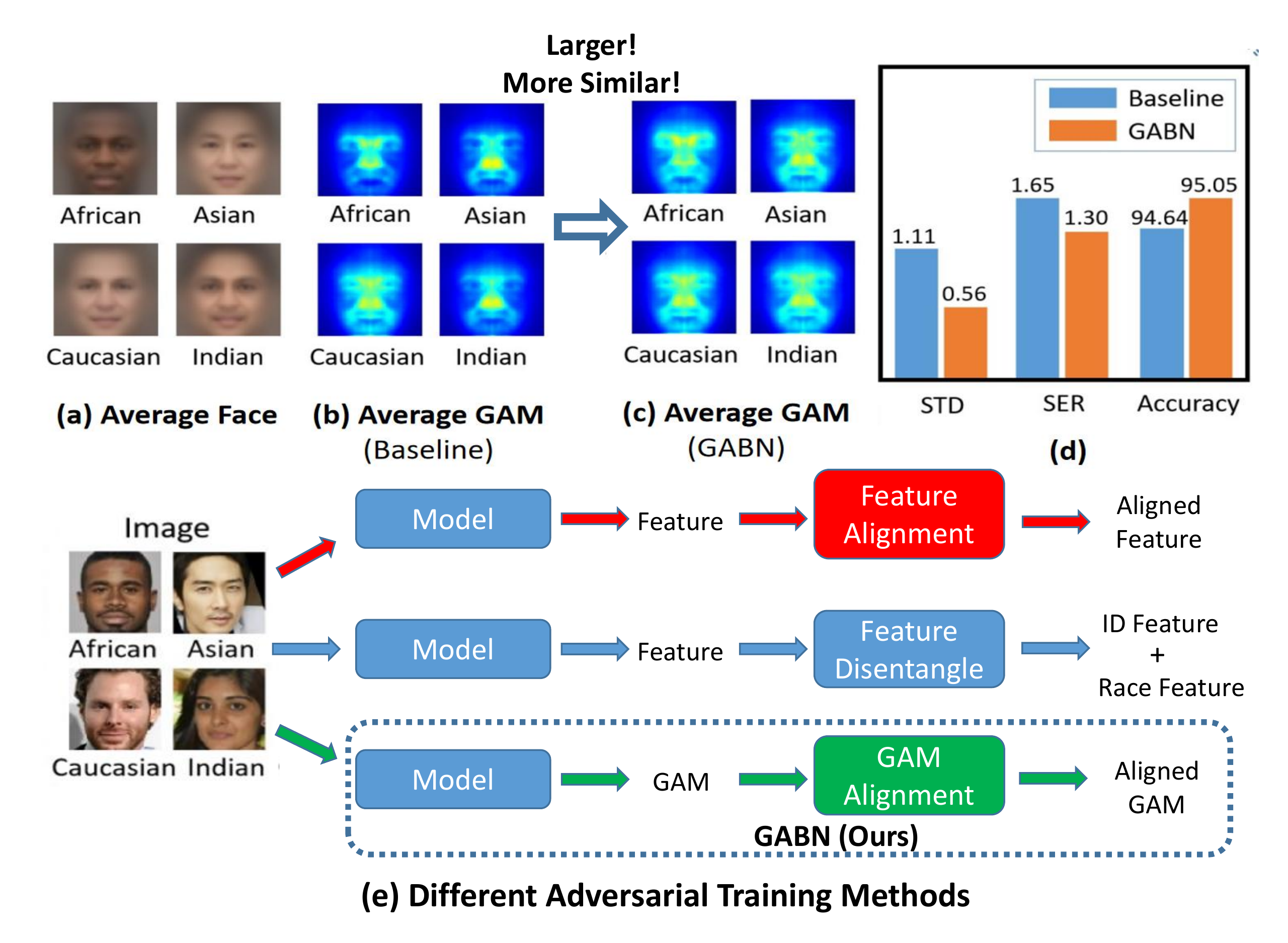}
\caption{
\textbf{(a)} Average faces of four races in the RFW dataset. 
\textbf{(b)} Average GAM generated by Baseline (ArcFace-34). 
\textbf{(c)} Average GAM generated by our Gradient Attention Balance Network (GABN). 
\textbf{(d)} Results on RFW dataset: We calculated standard deviation (STD) and skewed error ratio (SER) with the average accuracy of 4 races. 
STD and SER are used as the fairness criterion (specifically introduced in the SEC 4.1). 
Compared with the baseline of ArcFace-34 backbone, GABN improves the overall accuracy of face recognition and mitigates bias. GABN reduces STD from 1.11 to 0.56 and SER from 1.65 to 1.30. 
\textbf{(e)} Different adversarial training methods.
}
\label{fig:introduction}
\end{figure}

\section{Introduction}

With the development of deep learning \cite{hu2018squeeze,krizhevsky2012imagenet,simonyan2014very,szegedy2015going,wang2021deep,sun2014deep,schroff2015facenet}, the accuracy of the face recognition (FR) algorithm has reached a high level \cite{huang2008labeled,deng2019arcface,wang2018cosface}. 
However, with the emergence of more and more applications based on face recognition, its potential racial discrimination is attracting people's attention \cite{alvi2018turning,buolamwini2018gender}. 
For example, Amazon’s recognition tool incorrectly matched the photos of 28 U.S. congressmen with the photos of criminals (error rate up to 39\%). 
This unfair prediction leads to unfair treatment of different population groups. 
Therefore, it is very important to solve the problem of racial bias in the face recognition system.

Previous studies \cite{klare2012face,phillips2003face,dhar2020attributes,hill2019deep,terhorst2020beyond,yin2019feature,zhang2017range} have shown that the bias of face recognition comes from data and algorithms. 
Since the commonly used public large-scale datasets \cite{huang2008labeled,klare2015pushing,maze2018iarpa,phillips2012good,robinson2020face}, such as CASIA-WebFace \cite{yi2014learning}, VGGFace2 \cite{cao2018vggface2}, and MSCeleb-1M \cite{guo2016ms} are collected from the Internet, they inevitably encode race, gender, and age biases. 
Some studies \cite{wang2019racial,wang2020mitigate,wang2021meta,merler2019diversity} propose new and more balanced datasets. 
However, some studies \cite{wang2019racial,wang2020mitigate,wang2021meta,gong2020jointly,dhar2021pass,dhar2021distill} have proved that the racial bias of the models trained with such balanced datasets cannot be eliminated completely. 
Therefore, we also need to mitigate this bias through algorithms. Wang et al. \cite{wang2020mitigate,wang2021meta} propose the adaptive margin loss functions based on reinforcement-learning and meta-learning. 
Both methods use additional small fair datasets to guide the network to select the best margin for each race. 
Gong et al. \cite{gong2020jointly} propose a network that generates disentangled representations. 
This method mitigates bias by disentangling identity features and other demographic attribute features. 
They \cite{gong2021mitigating} also propose a method based on adaptive convolution kernel (GAC). 
GAC mitigates racial bias by introducing an additional group adaptive classifier. 
It is worth noting that the previous racial de-bias algorithms based on adversarial training mainly have two ways: 1) The method of feature alignment. 
2) The method of feature disentangle \cite{gong2020jointly, dhar2021pass}, as shown in Fig. \ref{fig:introduction} (e). 
Both of these methods delete racial features, resulting in the accuracy of face recognition lower than baseline. 
However, our method is to align the GAM of different races through adversarial training. 
In addition, our method not only improves fairness, but also improves model accuracy.

\begin{figure}[t]
\centering
\includegraphics[width=0.95\linewidth]{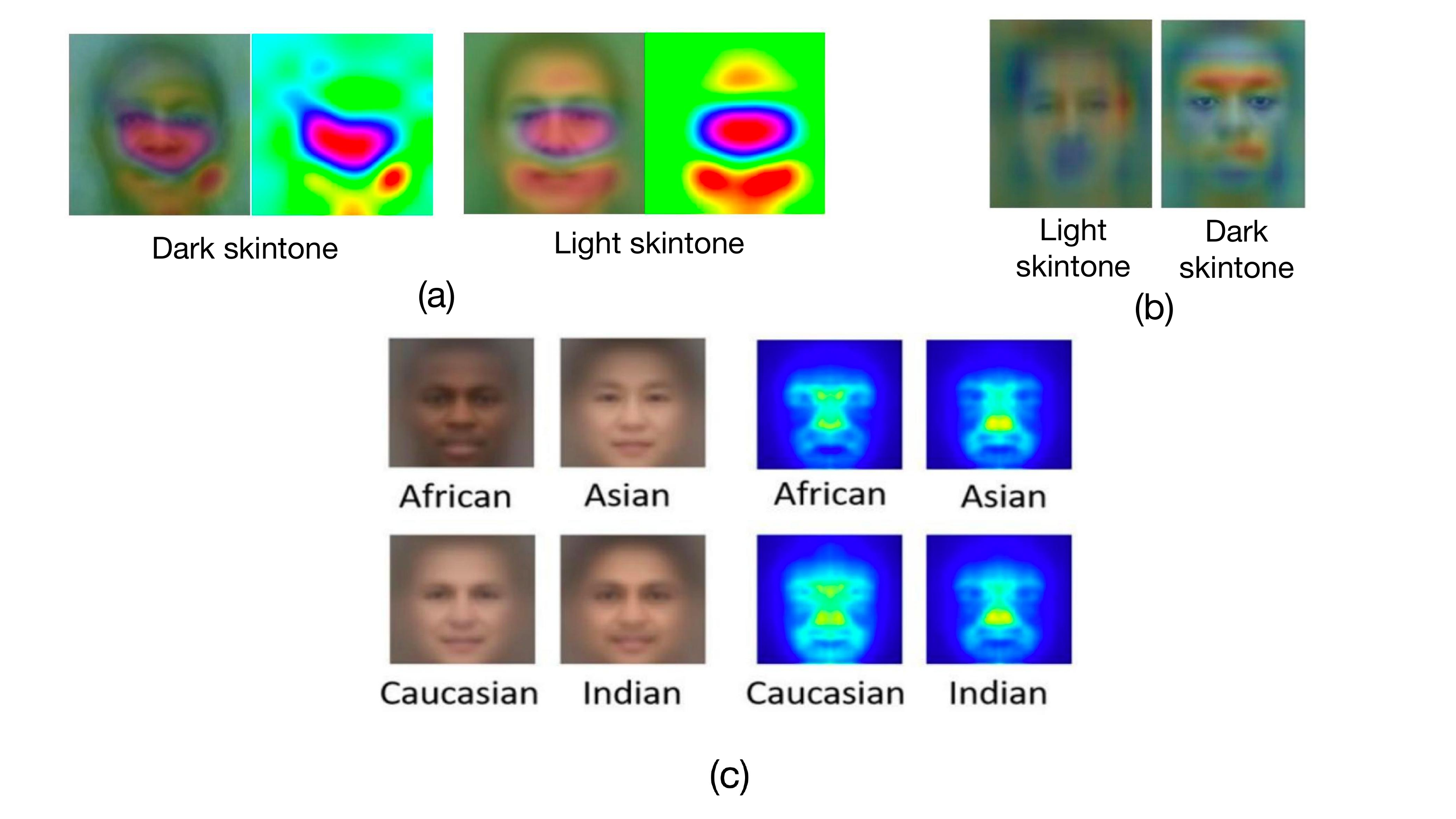}
\caption{
\textbf{(a)} This image is from \cite{dhar2021distill}. Average Gram-CAM for dark skintone and light skintone generated by Crystalface.
\textbf{(b)} This image is from \cite{nagpal2019deep}. Average Gram-CAM for dark skintone and light skintone generated by ArcFace.
\textbf{(c)} Ours. Average GAM generated by ArcFace. 
}
\label{fig:GAM}
\end{figure}

\begin{figure}[t]
\centering
\includegraphics[width=0.95\linewidth]{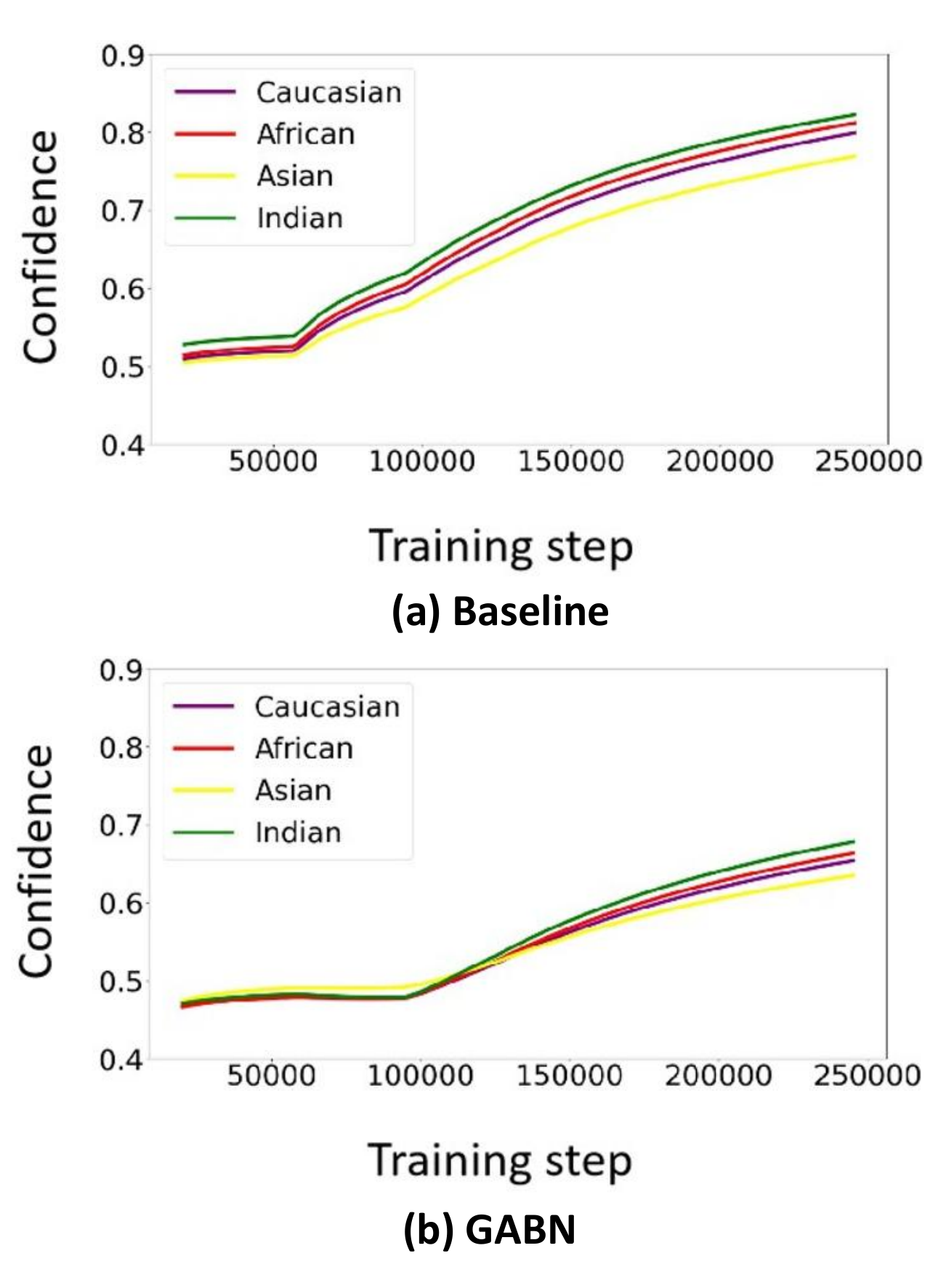}
\caption{
Comparison of confidence curves of four races (BUPT-Balancedface dataset). 
Confidence refers to the prediction probability value of the network for the category of the input image. 
(a) Confidence curve of Baseline (ArcFace-34). 
There is a gap in the confidence of different races. 
(b) Confidence curve of GABN. 
The gap between different races has narrowed.
}
\label{fig:confidence}

\end{figure}

\begin{figure*}[t]
\centering
\includegraphics[width=0.95\linewidth]{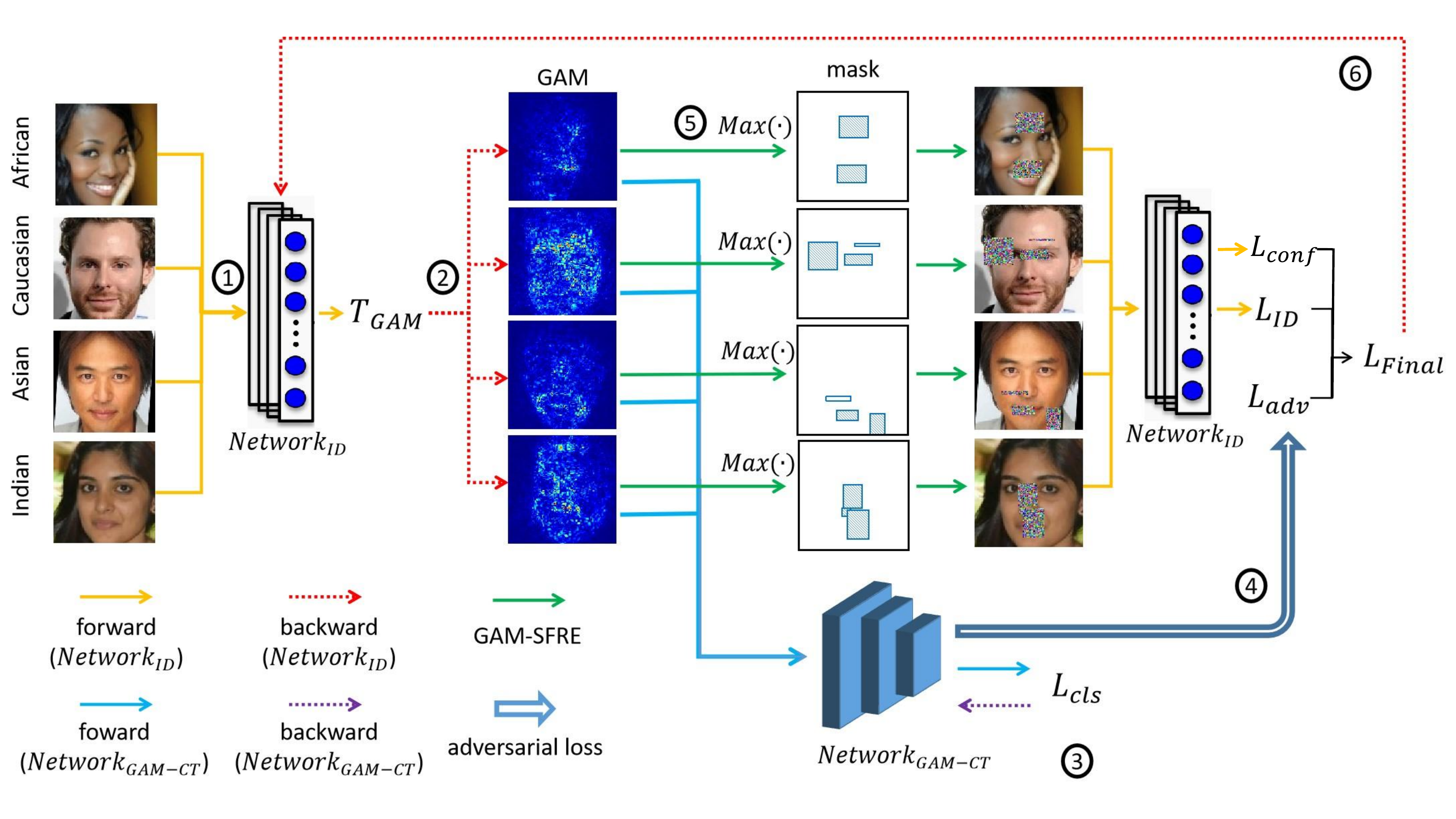}
\caption{
An illustration of our method. 
1) $Network_{ID}$ forward propagation to get $T_{GAM}$. 
2) $Network_{ID}$ backward propagation to obtain gradient attention map (GAM). 3) We input GAM into GAM race classification network ($Network_{GAM-CT}$) for training, so that the network can distinguish GAM of different races. 
4) We use $Network_{GAM-CT}$ as a discriminator. 
It provides an adversarial loss $L_{adv}$ for face recognition network ($Network_{ID}$). 
$L_{adv}$ will make $Network_{ID}$ generate GAM in which $Network_{GAM-CT}$ cannot distinguish race. 
5) According to GAM, calculate the Top-N sensitive regions, and use mask to erase the image. 
Then we input the erased image into the $Network_{ID}$ for training. 
6) We use $L_{conf}$, $L_{ID}$ and $L_{adv}$ for backward propagation and update the parameters of $Network_{ID}$.
}
\label{fig:GABN}
\end{figure*}

Some recent studies \cite{dhar2021distill,nagpal2019deep} have used Grad-cam \cite{selvaraju2017grad} for visualization and found that face recognition networks attend to different facial regions, depending on the race, as shown in Fig. \ref{fig:GAM}. 
For darker-skinned people, the face recognition network will pay more attention to a local region, such as the region near the nose for Africans. 
For Caucasians, the face recognition network focuses on a larger region. 
Inspired by Wang et al. \cite{wang2021representative}, we designed an image-leve map, called gradient attention map (GAM), as shown in Fig.~\ref{fig:GAM}. 
Grad-cam calculates the map based on the last convolution layer of the network and prefers to highlight the decision-making region. However, GAM can locate sensitive region more accurately. 
We used GAM for visualization and found a similar phenomenon: the sensitive region of darker-skinned people is smaller than that of Caucasians, and the sensitive region of darker-skinned people tends to be a local region, as shown in Fig.~\ref{fig:introduction} (b). Some studies \cite{wang2021meta} verified that darker-skinned people are more susceptible to noise and image quality than Caucasians. 
We think this is because face recognition networks pay more attention to local regions for darker-skinned people. 
This inconsistent sensitive region leads to racial bias in face recognition networks. 
In addition, we also found that during training, the network's confidence of different races is inconsistent, and this inconsistency will not completely disappear with the training, as shown in Fig.~\ref{fig:confidence} (a). Confidence refers to the prediction probability value of the network for the category of the input image. 
We believe that this inconsistency in confidence also introduces racial bias. To solve the above problems, we propose a new de-bias algorithm based on gradient attention, called Gradient Attention Balance Network (GABN). 
To solve the problem of inconsistent attention region, we propose two methods: 
1) GAM consistency training (GAM-CT): 
As shown in Fig.~\ref{fig:GABN}, we use the face recognition network ($Network_{ID}$) as a generator to generate the GAM of each image. 
In addition, we train a GAM race classification network ($Network_{GAM-CT}$) and use it as a discriminator. 
Then, we improve the consistency of GAM of different races through adversarial learning. 
2) GAM guided sensitive facial region erasure (GAM-SFRE): We try to use a special method to expand the sensitive region of the network on darker-skinned people and reduce the gap between darker-skinned people and Caucasians, as shown in Fig.~\ref{fig:GABN}. 
Under the guidance of GAM, we erase the Top-N sensitive regions in the image by generating masks, forcing the network to allocate attention to a larger facial region. 
In addition, to solve the problem of inconsistent confidence, we try to add confidence balance loss to the objective function. 
After adding the confidence balance loss, the confidence gap between the 4 races is reduced, as shown in Fig.~\ref{fig:confidence}(b). 
It is worth noting that our confidence balance loss is to obtain balanced confidence, not to maximize confidence.

A brief comparison between the baseline (ArcFace-34 \cite{deng2019arcface}) and our method (GABN) is sketched in Fig.~\ref{fig:introduction}. 
Comparing (b) and (c) in Fig.~\ref{fig:introduction}, we can see that our method enables the face recognition network to pay more consistent attention to different races, and successfully enables the face recognition network to allocate attention to a larger facial region. 
As shown in Fig.~\ref{fig:introduction}(d), our method not only enhances the overall accuracy but also greatly mitigates racial bias (STD, SER).

The main contributions of this work are as follows:

\begin{itemize}
	\item We propose a new face recognition de-bias algorithm called gradient attention balance network (GABN). 
 To our best knowledge, this is the first work to mitigate the racial bias of face recognition by improving the consistency of gradient attention map (GAM) of different races through adversarial learning (GAM-CT), which provides a new perspective to improve face recognition fairness. 
 Unlike previous work based on feature disentangle or feature alignment, our method is based on GAM alignment.
	\item Face recognition networks pay more attention to local regions for darker-skinned people. 
 To further reduce the performance gap between darker-skinned people and Caucasians, we use GAM guided sensitive facial region erasure (GAM-SFRE) to force the attention of the network to be allocated to a larger facial region. This method reduces the overfitting of darker-skinned people and improves the fairness of the network.
	\item Extensive experiments on the BUPT-Globalface \cite{wang2020mitigate}, BUPT-Balancedface \cite{wang2020mitigate}, and RFW \cite{wang2019racial} datasets show that our gradient attention balance network (GABN) shows more balanced performance.
\end{itemize}

\section{Related Work}

\textbf{Bias Mitigation in Face Recognition.} 
To mitigate the statistical bias of face recognition models, a variety of de-bias techniques \cite{gong2020jointly,gong2021mitigating,wang2019racial,wang2021meta,wang2020mitigate,dhar2021distill,dhar2021pass} have emerged. 
Wang et al. \cite{wang2019racial} proposed an information maximization adaptation network to mitigate the bias through domain adaptation. 
After that, Wang et al. \cite{wang2020mitigate,wang2021meta} proposed adaptive balance networks based on reinforcement-learning and meta-learning. 
They introduced a small and balanced verification set to guide the selection of margin parameters for each race during training. Xu et al. \cite{xu2021consistent} proposed a new loss function to mitigate the bias by improving the consistency of false positives. Gong et al. \cite{gong2020jointly} proposed using adversarial learning to disentangle the demographic attribute features and identity features. 
Recently, they \cite{gong2021mitigating} proposed a group adaptive classifier based on estimating demographic attributes to further improve the method. Unlike previous work, our GABN aims to make the network have similar gradient attention to people of different races. 
Mitigate racial prejudice by improving the consistency of gradient attention.

~\\
\noindent\textbf{De-bias Algorithm Based on Adversarial Learning (Face Recognition).}
Adversarial learning is widely used in the de-bias task of face recognition. Gong et al. \cite{gong2020jointly} proposed to use the disentangled representation for face recognition. 
They used an adversarial network with four classifiers, one for face recognition and the other three for demographic attributes. 
Adversarial learning is used to confuse and delete demographic information in features, so as to mitigate potential bias. 
Recent work \cite{dhar2021pass} proposed a new de-bias algorithm based on adversarial learning. 
They also disentangle the features of identity and demographic attributes through adversarial learning. The difference is that their work is based on the pre-trained model. 
Some work \cite{bortolato2020learning,li2019deepobfuscator,wu2018towards} has also used similar methods to eliminate sensitive information. 
Because demographic attribute features are an indispensable part of face identity, this method reduces the overall accuracy of face recognition. Different from previous work \cite{gong2020jointly,dhar2021pass}, our GABN uses adversarial learning to make the network have similar gradient attention to people of different races. 
GABN not only mitigates the bias but also improves the overall accuracy.

\section{Methodology}

\subsection{Problem Statement}
We visualized the images of 4 races to get the corresponding GAM. 
It is found that the sensitive region of darker-skinned people is small and that of Caucasians is large, as shown in Fig.~\ref{fig:introduction}(b). Fig.~\ref{fig:introduction}(b) is the average GAM of 4 races in the RFW dataset. 
We believe that this inconsistency in sensitive region introduces racial bias. In addition, when we trained the network, we found that the network's confidence in the 4 racial groups was inconsistent, as shown in Fig.~\ref{fig:confidence}(a). 
Therefore, we propose two methods to solve the problem of inconsistent attention region: 
1) We use GAM consistency training (GAM-CT) to improve the similarity of GAM. 2) We use GAM guided sensitive facial region erasure (GAM-SFRE) to force the GAM of darker-skinned people to tend to a larger facial region. 
We also propose to use confidence balance loss to solve the problem of inconsistent confidence. 
Specific details will be introduced in this section. 
The detail procedure of GABN is summarized in Algorithm \ref{alg1}.

\subsection{Gradient Attention Map (GAM)}
To know which facial regions are mainly concerned by the face recognition network ($Network_{ID}$), we use GAM to accurately locate the sensitive facial regions. 
As shown in Fig.~\ref{fig:GAM}, GAM is the image-level map. 
Each value of GAM accurately represents the sensitivity of $Network_{ID}$ to the corresponding pixels in the face image. 
As shown in Fig.~\ref{fig:GABN}(steps 1 and 2), forward propagation is used to obtain the probability value $P_i$ of each class (identity). 
Then, calculate the relative magnitude $T_{GAM}$ of the maximum probability value $P_{max}$ predicted by $Network_{ID}$ and the probability value $P_{i}$ of other classes. 
The formula is as follows:
\begin{equation}
    T_{GAM}=\frac{\sum_{i=1}^{N}(P_{max} - P_{i})}{N}
\end{equation}
where $N$ represents the number of identity categories. Because any perturbation will affect $P_i$ and $P_{max}$, we use $T_{GAM}$ as the objective function for backward propagation to obtain the gradient value and the GAM. 
The formula is as follows:
\begin{equation}
    GAM_{x,y} = \max(\nabla_{x,y}(abs(T_{GAM})))
\end{equation}
where $x$ and $y$ represent pixel coordinates, $\nabla$ represents the gradient value, the function $max(\cdot)$ calculates the maximum value along channel axis and the function $abs(\cdot)$ obtains the absolute value of each pixel. 
It is worth noting that Grad-cam \cite{selvaraju2017grad} is based on the last convolution layer of the network, which tends to highlight the decision-making region. 
GAM generates a map at the image-level to locate sensitive region. 
The positioning accuracy of GAM will be higher, but the attention distribution will be sparse, as shown in Fig.~\ref{fig:GAM}.

\subsection{GAM Consistency Training (GAM-CT)}
We use adversarial learning to improve the consistency of GAM among different races, as shown in Fig.~\ref{fig:GABN}(steps 3 and 4). 
Unlike previous works based on feature disentangle or feature alignment, our method is based on GAM alignment.
Firstly, a GAM race classification network ($Network_{GAM-CT}$) is trained to classify the races of GAM. 
The input of $Network_{GAM-CT}$ is the GAM of each image, and the output is the race corresponding to each GAM. 
$Network_{GAM-CT}$ consists of resnet-18 \cite{he2016deep} and fullly connected layer. 
The number of input channels of the first convolution layer of resnet-18 \cite{he2016deep} is changed to 1 (the dimension of GAM is W$\times$H$\times$1). 
The loss function $L_{cls}$ of $Network_{GAM-CT}$ is the standard cross entropy loss function:
\begin{equation}
    L_{cls} = -\frac{1}{N}\cdot\sum_{i=1}^{N}\log\frac{e^{W^T_{y_i}x_i+b_{y_i}}}{\sum_{j=1}^{n}e^{W^T_{j}x_i+b_{j}}}
\end{equation}
where $x_i$ denotes the deep feature of the $i_{th}$ sample, belonging to the $y_{i_{th}}$ class. 
$W_j$ denotes the $j_{th}$ column of the weight $W$ and $b_j$ is the bias term. 
The batch size and the class number are N and n.
Then, we use $Network_{GAM-CT}$ as discriminator and face recognition network ($Network_{ID}$) as generator for adversarial training. 
$Network_{GAM-CT}$ provides an adversarial loss $L_{adv}$ for face recognition network ($Network_{ID}$). 
$L_{adv}$ will make $Network_{ID}$ generate GAM in which $Network_{GAM-CT}$ cannot distinguish race. 
It is well known that a uniform distribution has the highest entropy and presents the most randomness. 
If an optimal classifier operating on GAM always produces a posterior probability of $\frac{1}{N}$ for all categories in the racial attribute, it means that the GAM generated by the $Network_{ID}$ has a consistent sensitive region of different races. 
$N$ represents the number of race. Therefore, we define the adversarial loss $L_{adv}$ as:
\begin{equation}
    L_{adv} = -\sum_{n=1}^{N}(\frac{1}{N}\cdot(\log\frac{e^{Network_{GAM-CT}(GAM)_n}}{\sum_{j=1}^{N}e^{Network_{GAM-CT}(GAM)_j}}))
\end{equation}
where $N$ is the number of races.
GAM-CT enables the face recognition network to pay more consistent attention to different races. 
See supplementary materials for specific algorithm process.

\subsection{GAM Guided Sensitive Facial Region Erasure (GAM-SFRE)}
To further improve fairness, we use GAM guided sensitive facial region erasure (GAM-SFRE) to expand the facial region of darker-skinned people, as shown in Fig.~\ref{fig:GABN} (step 5). 
First, we search each pixel of GAM and sort them in descending order according to the attention value of each pixel. 
Then, we select $N^{mask}$ pixels with the highest attention value as the central point $C^{mask}_{n}$ of the mask. 
The mask we use is a rectangular block with its height less than $h^{mask}$ and its width less than $w^{mask}$. 
Finally, we overlay the mask to the corresponding position of the original image, and then input the image into the network for training. 
See supplementary materials for specific algorithm process.

\begin{algorithm}[t]
\caption{GABN (GAM-CT + GAM-SFRE)}
\label{alg1}
\textbf{Require:}\\
Original facial images $I_O$.\\
Gradient attention map $GAM$.\\
Max number of epochs $E$.\\
Number of network updates per epoch $B$.\\
Mask count $N$.\\
\textbf{Ensure:}\\
Face recognition network $Network_{ID}$.\\
GAM race classification network $Network_{GAM-CT}$.

\textbf{For} {$e$ \textbf{in} $E$}

\setlength{\parindent}{1em} \textbf{For} {$b$ \textbf{in} $B$} 

        \setlength{\parindent}{2em} FP $Network_{ID}$ to obtain $P_{max}$ and $P_i$;
        
        Use $P_{max}$ to get $L_{conf}$;
        
        Use formula 1 in the paper to get $T_{GAM}$;
        
        BP $Network_{ID}$ to obtain $GAM$ (formula 2);
        
        Get the erased image $I_E$ guided by $GAM$:
        
        n = 0;

        \setlength{\parindent}{2em} \textbf{For} {$n$ \textbf{in} $N$}
            
            \setlength{\parindent}{3em} $i, j$ = coordinate of the n$th$ largest value; 
            
            $H_t$ = Rand(1, $H^{mask}$);
            
            $W_t$ = Rand(1, $W^{mask}$);
            
            Use a rectangle with $H_t$ and $W_t$ to erase $I_O$;
        
       \setlength{\parindent}{2em} Training $Network_{GAM-CT}$; 
        
        BP to optimize $Network_{GAM-CT}$ (formula 3);
        
        Training $Network_{ID}$;
        
        FP $Network_{GAM-CT}$ to get $L_{adv}$ (formula 4);
        
        Use $L_{id}$, $L_{conf}$ and $L_{adv}$ to get $L_{final}$;
        
        Use $L_{final}$ to optimize $Network_{ID}$.

\end{algorithm}

\subsection{Objective Function}

As described in SEC 3.1, we believe that this inconsistency in confidence is also one of the factors leading to racial bias. 
Therefore, based on the objective function of ArcFace \cite{deng2019arcface}, we add a confidence balance loss to reduce this inconsistency. 
The details are as follows: 
1) The normalized features and weights are inputted into the loss function of ArcFace (additive angular margin penalty) to get the probability value $P_i$ of each class, and get the maximum probability value $P_{max}$ from it, which is called confidence. 
2) Calculate the number of samples with $P_{max}$ less than the threshold $T_{confidence}$ in the current batch ($n_{confidence}$). 
3) Get confidence balanced loss function:
\begin{equation}
    L_{conf} = \frac{n_{batch}-n_{confidence}}{n_{batch}}
\end{equation}
where $n_{batch}$ represents the number of samples in current batch.
4) Finally, we combine $L_{conf}$, $L_{ID}$ and adversarial loss $L_{adv}$ in GAM-CT to obtain the final objective function $L_{Final}$. 
Here, $L_{conf}$ and $L_{adv}$ are taken as the penalty coefficient.
\begin{equation}
    L_{ID} = 
    -\frac{1}{N}\sum_{i=1}^{N}\log\frac{e^{s(\cos(\theta_{y_i}+m))}}{e^{s(\cos(\theta_{y_i}+m))}+\sum\limits_{j=1}^{n}e^{s(\cos(\theta_j))}}
\end{equation}

\begin{equation}
    K = L_{conf} + L_{adv}
\end{equation}

\begin{equation}
    L_{Final} = 
    -\frac{1}{N}\sum_{i=1}^{N}\log\frac{e^{s(\cos(\theta_{y_i}+m))}}{e^{s(\cos(\theta_{y_i}+m))}+\sum\limits_{j=1}^{n}e^{s(\cos(\theta_j+K))}}
\end{equation}

 We update the parameters of face recognition network ($Network_{ID}$) by minimizing $L_{Final}$.

\begin{table*}[t]
	\begin{center}
    \normalsize
    
    \caption{Verification performance (\%) of protocol on RFW with SOTA methods ([BUPT-Globalface, ResNet34, Loss-ArcFace]).} 
    \label{tab:BUPT_Globalface}
    {
	\begin{tabular}{c|cccc|c|cc}
		\hline
         Methods & Caucasian & Indian & Asian & African & Avg & \multicolumn{2}{c}{Fairness} \\
         & & & & & & STD & SER \\ \hline \hline
         Baseline (ArcFace) \cite{deng2019arcface} & 97.37 & 95.68 & 94.55 & 93.87 & 95.37 & 1.53 & 2.33 \\
         
         M-RBN \cite{wang2020mitigate} & 97.03 & 95.58 & 94.40 & 95.18 & 95.55 & 1.10 & 1.89 \\

         RL-RBN \cite{wang2020mitigate} & 97.08 & 95.63 & 95.57 & 94.87 & 95.79 & 0.93 & 1.76 \\
         
         MBN \cite{wang2021meta} & 96.87 & 96.20 & 95.63 & 95.00 & 95.93 & 0.80 & 1.60 \\
         \textbf{GABN (Ours)} & 97.21 & 96.06 & 95.73 & 95.51 & 96.13 & \textbf{0.75} & \textbf{1.60} \\ \hline
	\end{tabular}}
    \end{center}
\end{table*}

\begin{table*}[t]
	\begin{center}
    \normalsize
    
    \caption{Verification performance (\%) of protocol on RFW with SOTA methods ([BUPT-Balancedface, ResNet34, Loss-ArcFace]).} 
    \label{tab:BUPT_Balancedface}
    
    {
	\begin{tabular}{c|cccc|c|cc}
		\hline
         Methods & Caucasian & Indian & Asian & African & Avg & \multicolumn{2}{c}{Fairness} \\
         & & & & & & STD & SER \\ \hline \hline
         Baseline (ArcFace) \cite{deng2019arcface} & 96.18 & 94.67 & 93.72 & 93.98 & 94.64 & 1.11 & 1.65 \\
         
         ACNN \cite{kang2017incorporating} & 96.12 & 94.55 & 93.67 & 94.00 & 94.58 & 1.08 & 1.63 \\
         
         PFE \cite{shi2019probabilistic} & 96.38 & 94.60 & 94.27 & 95.17 & 95.11 & 0.93 & 1.58 \\
         
         DebFace \cite{gong2020jointly} & 95.95 & 94.78 & 94.33 & 93.67 & 94.68 & 0.96 & 1.56 \\
         
         GAC \cite{gong2021mitigating} & 96.02 & 94.22 & 94.10 & 94.12 & 94.62 & 0.93 & 1.48 \\
         
         RL-RBN \cite{wang2020mitigate} & 96.27 & 94.68 & 94.82 & 95.00 & 94.64 & 0.73 & 1.43 \\
         
         MBN \cite{wang2021meta}  & 96.25 & 95.32 & 94.85 & 95.38 & 95.45 & 0.58 & 1.37 \\
         \textbf{GABN (Ours)} & 95.78 & 95.21 & 94.51 & 94.71 & 95.05 & \textbf{0.56} & \textbf{1.30} \\ \hline
	\end{tabular}}
    \end{center}
    
\end{table*}

\section{Experiments}

\subsection{Experimental Setting}
\textbf{Dataset:} Our bias study uses BUPT-Balancedface \cite{wang2020mitigate} and BUPT-Globalface datasets \cite{wang2020mitigate} for training and RFW \cite{wang2019racial} datasets for testing. 
BUPT-Balancedface consists of faces of 4 races: Caucasians, Asians, Indians, and Africans. 
This dataset contains 1.3 million images of 28K celebrities, with about 7K identities per race. 
BUPT-Globalface contains 2 million images of 38K celebrities, whose racial distribution is roughly the same as the real distribution of the world's population. 
RFW consists of faces of 4 races: Caucasians, Asians, Indians and Africans. Each subset of RFW contains about 10K images of 3K identities.

\textbf{Evaluation Protocol.}
The evaluation protocol we used are consistent with the previous work\cite{wang2020mitigate,wang2021meta,gong2020jointly,gong2021mitigating}. 
We calculated standard deviation (STD) and skewed error ratio (SER) with the average accuracy of 4 races. 
STD and SER are used as the fairness criterion. 
STD reflects the amount of dispersion of accuracies of different races. 
SER is computed by the ratio of the highest error rate to the lowest error rate among 4 races. 
 $SER = \frac{max Error_g}{min Error_g}$,
where g in $Caucasian, Indian, Asian, African$.

\textbf{Implementation details:} For preprocessing, we use 5 facial landmarks for similarity transformation, then crop and resize the images to 112 × 112 pixels. 
Each pixel ([0, 255]) in RGB images are normalized by subtracting 127.5 and then being divided by 128. 
Consistent with the previous work \cite{wang2020mitigate,wang2021meta,gong2020jointly,gong2021mitigating,xu2021consistent}, the CNN architecture we use is ResNet-34 \cite{he2016deep}. 
The CNN model is trained on 3 GPUs (NVIDIA GeForce 1080Ti). 
The models are trained by the SGD algorithm, with momentum 0.9 and weight decay 5e-4. 
We set the scale and margin as $s$ = 64 and $m$ = 0.35 according to the common settings in the previous work \cite{xu2021consistent}. 
The confidence threshold $T_{confidence}$ is set to 70\%. 
On BUPT-Balancedface, the batch size is set to 200, the learning rate starts from 0.1 and is divided by 10 at 9, 15, 20 epochs. 
The training process is finished at 30 epochs. 
On BUPT-Globalface, the batch size is set to 250, we divide the learning rate at 9, 15, 20 epochs and finish at 40 epochs. 
Our GAM race classifier uses ResNet-18 \cite{he2016deep}, in which the number of input channels of the first convolution layer of ResNet-18 is changed to 1. The GAM race classifier is trained by the SGD algorithm, with momentum 0.9 and weight decay 5e-4. 
Its learning rate starts from 0.1 and is divided by 10 at 9, 15, 20 epochs. 
It should be noted that the face recognition network and GAM race classifier are trained in turn.

\subsection{Comparisons with SOTA Methods}
Consistent with the previous work\cite{kang2017incorporating, wang2020mitigate, gong2021mitigating, wang2021meta, shi2019probabilistic, gong2020jointly}, we train a ResNet-34 model on BUPT-Globalface, use the loss function of ArcFace \cite{deng2019arcface}, and then report the results on RFW protocol, as shown in Table~\ref{tab:BUPT_Globalface}. 
Compared with the SOTA results, GABN reduces STD to 0.75 and SER to 1.60. 
We also used the same method to conduct experiments on BUPT-Balancedface, as shown in Table~\ref{tab:BUPT_Balancedface}. 
Compared with the SOTA results, GABN reduces STD to 0.56. 
Compared with DebFace \cite{gong2020jointly}, which is also based on adversarial learning, our method not only obtains lower STD and SER but also improves the overall accuracy of the model. 
The above results show that our method has achieved competitive performance on both balanced and unbalanced datasets.

\begin{table*}[t]
	\begin{center}
    \normalsize 
    
    \caption{Verification performance (\%) of different threshold of confidence balanced loss ($T_{confidence}$).} 
    
    \label{tab:confidence}
    
    {
	\begin{tabular}{c|ccccc|c|cc}
		\hline
         Methods& $T_{confidence}$ & Caucasian & Indian & Asian & African & Avg & \multicolumn{2}{c}{Fairness} \\
         & & & & & & & STD & SER \\ \hline \hline
         Baseline & - & 96.18 & 94.67 & 93.72 & 93.98 & 94.64 & 1.11 & 1.65 \\ \hline

         &60\%  & 95.21 & 94.65 & 93.73 & 94.05 & 94.44 & 0.66 & \textbf{1.31} \\
         
         + confidence & 70\%  & 96.23 & 95.41 & 94.66 & 95.15 & 95.36 & \textbf{0.65} & 1.42 \\
         
         balanced loss & 75\%  & 96.13 & 95.06 & 94.45 & 94.83 & 95.12 & 0.72 & 1.43 \\
         
         &80\%  & 95.61 & 94.36 & 93.58 & 93.88 & 94.36 & 0.89 & 1.46 \\
     \hline
         
	\end{tabular}}
    \end{center}
    
\end{table*}

\begin{table*}[t]
	\begin{center}
    \normalsize
    
    \caption{Ablation of different components on the RFW protocol. \textbf{+Random erase} is the result of using random erase (RE).} 
    
    \label{tab:different_components}
    
    {
	\begin{tabular}{c|cccc|c|cc}
		\hline
         Methods & Caucasian & Indian & Asian & African & Avg & \multicolumn{2}{c}{Fairness} \\
         & & & & & & STD & SER \\ \hline \hline
         Baseline (ArcFace-34) & 96.18 & 94.67 & 93.72 & 93.98 & 94.64 & 1.11 & 1.65 \\
         
         + Random Erase & 96.20 & 95.40 & 94.57 & 94.57 & 95.18 & 0.78 & 1.43 \\

         + confidence balance loss & 96.23 & 95.41 & 94.66 & 95.15 & 95.36 & 0.65 & 1.42 \\
         
         GAM-CT & 95.93 & 95.20 & 94.41 & 94.90 & 95.11 & 0.63 & 1.37 \\
         
         GAM-SFRE & 96.16 & 95.43 & 94.63 & 95.08 & 95.32 & 0.64 & 1.39 \\
         
         \textbf{GABN (GAM)} & 95.78 & 95.21 & 94.51 & 94.71 & 95.05 & \textbf{0.56} & \textbf{1.30} \\ \hline
	\end{tabular}}
    \end{center}

\end{table*}

\subsection{Ablation Study}
To investigate the efficacy of our method, we conduct a lot of ablation studies. 
The setup of ablation study is consistent with that described in SEC 4.1. 
We use the framework of ArcFace-34, then train Baseline and GABN on BUPT-Balancedface dataset, and test on RFW dataset.

\textbf{Effect of the confidence balanced loss.} 
We use $T_{confidence}$ to control the confidence threshold. Table~\ref{tab:confidence} reports the results of different $T_{confidence}$. $T_{confidence}$ is not the higher the better. 
The best results are obtained when we set $T_{confidence}$ to 70\%. 
This is in line with our expectations, because our confidence balanced loss is to obtain a more balanced confidence, not to maximize the confidence. 
We use this setting in later experiment.

\textbf{Effect of the GAM-SFRE.} 
To study the effectiveness of GAM-SFRE module, We compare our GAM-SFRE with random erasure (RE). 
We keep the settings of RE and GAM-SFRE consistent and conduct experiments. The result of our method is better than random erasure, as shown in Table~\ref{tab:different_components}. 
We explain the reasons as follows: Although random erasure increases the diversity of samples, most of the random erasure regions are insensitive, which makes the performance improvement of the network limited. 
However, our method is based on GAM guidance for erasure. 
Each erased region is a sensitive region, forcing the network to focus on a larger facial region, so as to effectively reduce overfitting. 
We also compared RE and GAM-SFRE through visualization, as shown in SEC 4.4.

\textbf{Effect of the GAM-CT.} 
To study the effectiveness of GAM-CT module, we use GAM-CT for ablation study, as shown in Table. \ref{tab:different_components}. 
When we used GAM-CT, the STD decreased to 0.63.

We can get better results than Baseline when we use GAM-CT or GAM-SFRE. 
This shows that these components can effectively improve the fairness of the model. 
When we use these modules at the same time, we can get the performance of SOTA. 
We summarize the best results of each module in Table~\ref{tab:different_components}.

\begin{figure}[t]
\centering
\includegraphics[width=0.95\linewidth]{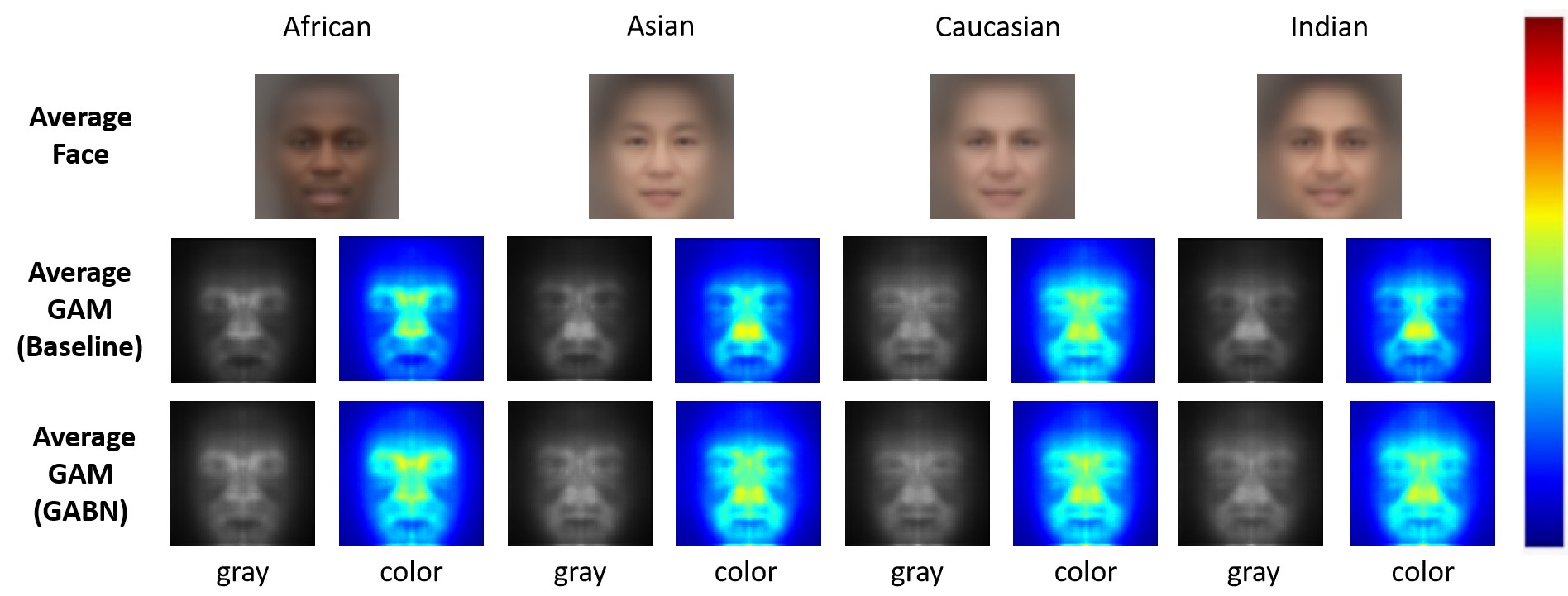}
\caption{
(a) Comparison of average GAM of 4 races of Baseline and GABN. 
We convert GAM from gray to color. Red represents greater attention. 
}
\label{fig:visual_average}
\end{figure}

\subsection{Visualization and Analysis}
\begin{figure}[t]
\centering
\includegraphics[width=0.95\linewidth]{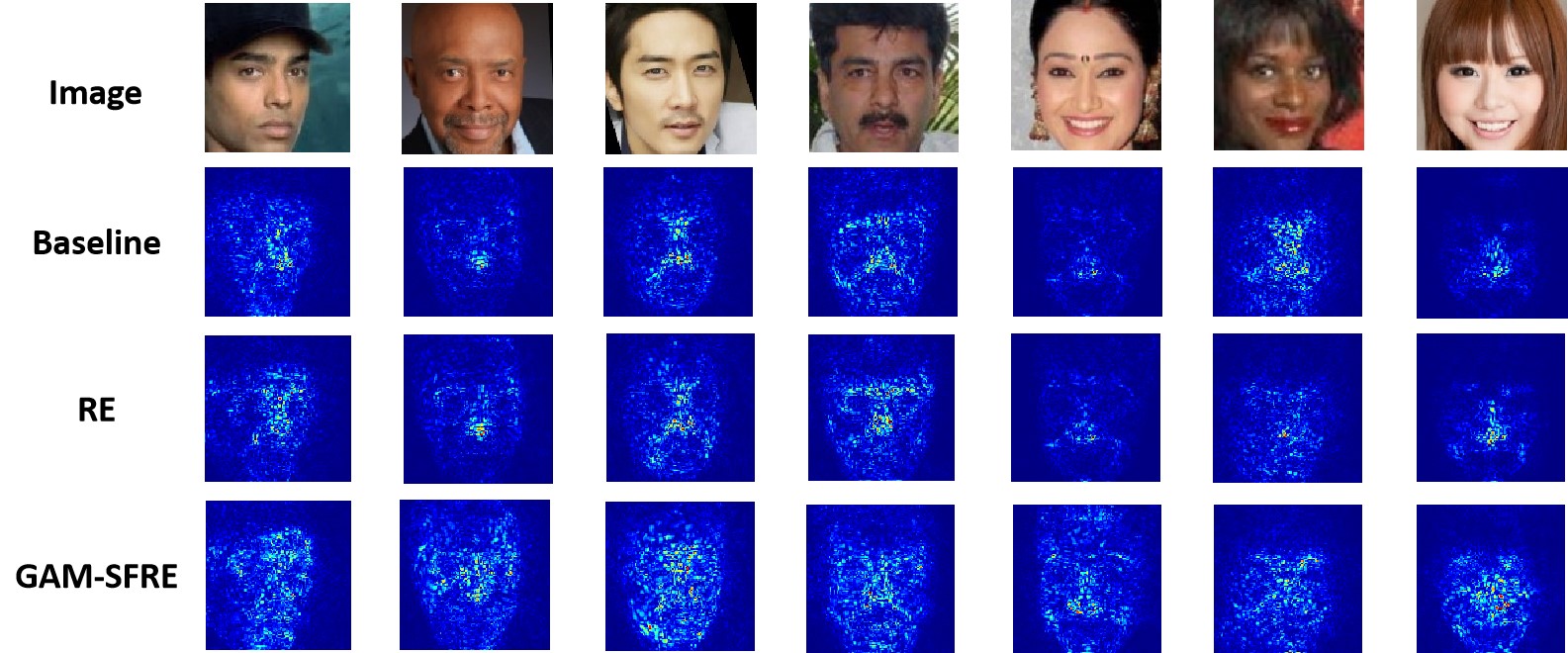}
\caption{
GAM comparison diagram of random erase (RE) and GAM-SFRE.
}
\label{fig:visual_GAMSFRE}
\end{figure}

To prove that our method improves the consistency of GAM of different races, we visualize GAM. 
We use the method introduced in SEC 3 to obtain the GAM of each image in the RFW dataset and calculate the average GAM of 4 races. 
As shown in Fig.~\ref{fig:visual_average}, in the average GAM of the 4 races obtained by Baseline, the sensitive region of Caucasians is large and that of darker-skinned people is small. 
However, the sensitive region of GABN has higher consistency and larger sensitive region. 
This shows that our GABN has successfully reduced the attention gap between different races. 
In addition, we also visualize the GAM of random erasure (RE) and GAM-SFRE, as shown in Fig.~\ref{fig:visual_GAMSFRE}. 
RE can expand sensitive region, but the effect is very small. 
GAM-SFRE can expand the region to the whole face, and the effect is remarkable.

\subsection{Limitation Analysis}
As shown in Fig.~\ref{fig:GABN}, GABN needs 2 backward propagation to train 1 step. 
The first backward propagation generates GAM, and the second backward propagation updates the parameters. 
In addition, we need to train a GAM race classification network ($Network_{GAM-CT}$). 
We train on 3 GPUs (NVIDIA Geforce 1080TI), which takes about 2.2 times longer than baseline (ArcFace-34). 
However, we think the training time is within the acceptable range.

\section{Conclusion}
In this paper, we propose a de-bias network (GABN) based on gradient attention and adversarial learning to reduce racial bias. 
GABN improves the consistency of gradient attention map (GAM) of different races through adversarial learning. 
In addition, GABN also uses GAM guided sensitive region erasure to expand the sensitive region of darker-skinned people, further reducing the performance gap between darker-skinned people and Caucasians. 
Comprehensive experiments show the effectiveness of our GABN.

\noindent\textbf{Acknowledgements.}
This work was supported by China Postdoctoral Science Foundation under Grant 2022M720517.

{\small
\bibliographystyle{ieee_fullname}
\bibliography{egbib}

\begin{thebibliography}{10}\itemsep=-1pt

\bibitem{alvi2018turning}
Mohsan Alvi, Andrew Zisserman, and Christoffer Nell{\aa}ker.
\newblock Turning a blind eye: Explicit removal of biases and variation from
  deep neural network embeddings.
\newblock In {\em Proceedings of the European Conference on Computer Vision
  (ECCV) Workshops}, pages 0--0, 2018.

\bibitem{bortolato2020learning}
Bla{\v{z}} Bortolato, Marija Ivanovska, Peter Rot, Janez Kri{\v{z}}aj, Philipp
  Terh{\"o}rst, Naser Damer, Peter Peer, and Vitomir {\v{S}}truc.
\newblock Learning privacy-enhancing face representations through feature
  disentanglement.
\newblock In {\em 2020 15th IEEE International Conference on Automatic Face and
  Gesture Recognition (FG 2020)}, pages 495--502. IEEE, 2020.

\bibitem{buolamwini2018gender}
Joy Buolamwini and Timnit Gebru.
\newblock Gender shades: Intersectional accuracy disparities in commercial
  gender classification.
\newblock In {\em Conference on fairness, accountability and transparency},
  pages 77--91. PMLR, 2018.

\bibitem{cao2018vggface2}
Qiong Cao, Li Shen, Weidi Xie, Omkar~M Parkhi, and Andrew Zisserman.
\newblock Vggface2: A dataset for recognising faces across pose and age.
\newblock In {\em 2018 13th IEEE international conference on automatic face \&
  gesture recognition (FG 2018)}, pages 67--74. IEEE, 2018.

\bibitem{deng2019arcface}
Jiankang Deng, Jia Guo, Niannan Xue, and Stefanos Zafeiriou.
\newblock Arcface: Additive angular margin loss for deep face recognition.
\newblock In {\em Proceedings of the IEEE/CVF conference on computer vision and
  pattern recognition}, pages 4690--4699, 2019.

\bibitem{dhar2020attributes}
Prithviraj Dhar, Ankan Bansal, Carlos~D Castillo, Joshua Gleason, P~Jonathon
  Phillips, and Rama Chellappa.
\newblock How are attributes expressed in face dcnns?
\newblock In {\em 2020 15th IEEE International Conference on Automatic Face and
  Gesture Recognition (FG 2020)}, pages 85--92. IEEE, 2020.

\bibitem{dhar2021pass}
Prithviraj Dhar, Joshua Gleason, Aniket Roy, Carlos~D Castillo, and Rama
  Chellappa.
\newblock Pass: Protected attribute suppression system for mitigating bias in
  face recognition.
\newblock In {\em Proceedings of the IEEE/CVF International Conference on
  Computer Vision}, pages 15087--15096, 2021.

\bibitem{dhar2021distill}
Prithviraj Dhar, Joshua Gleason, Aniket Roy, Carlos~D Castillo, P~Jonathon
  Phillips, and Rama Chellappa.
\newblock Distill and de-bias: Mitigating bias in face recognition using
  knowledge distillation.
\newblock {\em arXiv preprint arXiv:2112.09786}, 2021.

\bibitem{gong2020jointly}
Sixue Gong, Xiaoming Liu, and Anil~K Jain.
\newblock Jointly de-biasing face recognition and demographic attribute
  estimation.
\newblock In {\em European conference on computer vision}, pages 330--347.
  Springer, 2020.

\bibitem{gong2021mitigating}
Sixue Gong, Xiaoming Liu, and Anil~K Jain.
\newblock Mitigating face recognition bias via group adaptive classifier.
\newblock In {\em Proceedings of the IEEE/CVF Conference on Computer Vision and
  Pattern Recognition}, pages 3414--3424, 2021.

\bibitem{guo2016ms}
Yandong Guo, Lei Zhang, Yuxiao Hu, Xiaodong He, and Jianfeng Gao.
\newblock Ms-celeb-1m: A dataset and benchmark for large-scale face
  recognition.
\newblock In {\em European conference on computer vision}, pages 87--102.
  Springer, 2016.

\bibitem{he2016deep}
Kaiming He, Xiangyu Zhang, Shaoqing Ren, and Jian Sun.
\newblock Deep residual learning for image recognition.
\newblock In {\em Proceedings of the IEEE conference on computer vision and
  pattern recognition}, pages 770--778, 2016.

\bibitem{hill2019deep}
Matthew~Q Hill, Connor~J Parde, Carlos~D Castillo, Y~Ivette Colon, Rajeev
  Ranjan, Jun-Cheng Chen, Volker Blanz, and Alice~J O’Toole.
\newblock Deep convolutional neural networks in the face of caricature.
\newblock {\em Nature Machine Intelligence}, 1(11):522--529, 2019.

\bibitem{hu2018squeeze}
Jie Hu, Li Shen, and Gang Sun.
\newblock Squeeze-and-excitation networks.
\newblock In {\em Proceedings of the IEEE conference on computer vision and
  pattern recognition}, pages 7132--7141, 2018.

\bibitem{huang2008labeled}
Gary~B Huang, Marwan Mattar, Tamara Berg, and Eric Learned-Miller.
\newblock Labeled faces in the wild: A database forstudying face recognition in
  unconstrained environments.
\newblock In {\em Workshop on faces in'Real-Life'Images: detection, alignment,
  and recognition}, 2008.

\bibitem{kang2017incorporating}
Di Kang, Debarun Dhar, and Antoni Chan.
\newblock Incorporating side information by adaptive convolution.
\newblock {\em Advances in Neural Information Processing Systems}, 30, 2017.

\bibitem{klare2012face}
Brendan~F Klare, Mark~J Burge, Joshua~C Klontz, Richard W~Vorder Bruegge, and
  Anil~K Jain.
\newblock Face recognition performance: Role of demographic information.
\newblock {\em IEEE Transactions on Information Forensics and Security},
  7(6):1789--1801, 2012.

\bibitem{klare2015pushing}
Brendan~F Klare, Ben Klein, Emma Taborsky, Austin Blanton, Jordan Cheney,
  Kristen Allen, Patrick Grother, Alan Mah, and Anil~K Jain.
\newblock Pushing the frontiers of unconstrained face detection and
  recognition: Iarpa janus benchmark a.
\newblock In {\em Proceedings of the IEEE conference on computer vision and
  pattern recognition}, pages 1931--1939, 2015.

\bibitem{krizhevsky2012imagenet}
Alex Krizhevsky, Ilya Sutskever, and Geoffrey~E Hinton.
\newblock Imagenet classification with deep convolutional neural networks.
\newblock {\em Advances in neural information processing systems}, 25, 2012.

\bibitem{li2019deepobfuscator}
Ang Li, Jiayi Guo, Huanrui Yang, and Yiran Chen.
\newblock Deepobfuscator: Adversarial training framework for privacy-preserving
  image classification.
\newblock {\em arXiv preprint arXiv:1909.04126}, 2(3), 2019.

\bibitem{maze2018iarpa}
Brianna Maze, Jocelyn Adams, James~A Duncan, Nathan Kalka, Tim Miller, Charles
  Otto, Anil~K Jain, W~Tyler Niggel, Janet Anderson, Jordan Cheney, et~al.
\newblock Iarpa janus benchmark-c: Face dataset and protocol.
\newblock In {\em 2018 international conference on biometrics (ICB)}, pages
  158--165. IEEE, 2018.

\bibitem{merler2019diversity}
Michele Merler, Nalini Ratha, Rogerio~S Feris, and John~R Smith.
\newblock Diversity in faces.
\newblock {\em arXiv preprint arXiv:1901.10436}, 2019.

\bibitem{nagpal2019deep}
Shruti Nagpal, Maneet Singh, Richa Singh, and Mayank Vatsa.
\newblock Deep learning for face recognition: Pride or prejudiced?
\newblock {\em arXiv preprint arXiv:1904.01219}, 2019.

\bibitem{phillips2012good}
P~Jonathon Phillips, J~Ross Beveridge, Bruce~A Draper, Geof Givens, Alice~J
  O'Toole, David Bolme, Joseph Dunlop, Yui~Man Lui, Hassan Sahibzada, and
  Samuel Weimer.
\newblock The good, the bad, and the ugly face challenge problem.
\newblock {\em Image and Vision Computing}, 30(3):177--185, 2012.

\bibitem{phillips2003face}
P~Jonathon Phillips, Patrick Grother, Ross Micheals, Duane~M Blackburn, Elham
  Tabassi, and Mike Bone.
\newblock Face recognition vendor test 2002.
\newblock In {\em 2003 IEEE International SOI Conference. Proceedings (Cat. No.
  03CH37443)}, page~44. IEEE, 2003.

\bibitem{robinson2020face}
Joseph~P Robinson, Gennady Livitz, Yann Henon, Can Qin, Yun Fu, and Samson
  Timoner.
\newblock Face recognition: too bias, or not too bias?
\newblock In {\em Proceedings of the ieee/cvf conference on computer vision and
  pattern recognition workshops}, pages 0--1, 2020.

\bibitem{schroff2015facenet}
Florian Schroff, Dmitry Kalenichenko, and James Philbin.
\newblock Facenet: A unified embedding for face recognition and clustering.
\newblock In {\em Proceedings of the IEEE conference on computer vision and
  pattern recognition}, pages 815--823, 2015.

\bibitem{selvaraju2017grad}
Ramprasaath~R Selvaraju, Michael Cogswell, Abhishek Das, Ramakrishna Vedantam,
  Devi Parikh, and Dhruv Batra.
\newblock Grad-cam: Visual explanations from deep networks via gradient-based
  localization.
\newblock In {\em Proceedings of the IEEE international conference on computer
  vision}, pages 618--626, 2017.

\bibitem{shi2019probabilistic}
Yichun Shi and Anil~K Jain.
\newblock Probabilistic face embeddings.
\newblock In {\em Proceedings of the IEEE/CVF International Conference on
  Computer Vision}, pages 6902--6911, 2019.

\bibitem{simonyan2014very}
Karen Simonyan and Andrew Zisserman.
\newblock Very deep convolutional networks for large-scale image recognition.
\newblock {\em arXiv preprint arXiv:1409.1556}, 2014.

\bibitem{sun2014deep}
Yi Sun, Yuheng Chen, Xiaogang Wang, and Xiaoou Tang.
\newblock Deep learning face representation by joint
  identification-verification.
\newblock {\em Advances in neural information processing systems}, 27, 2014.

\bibitem{szegedy2015going}
Christian Szegedy, Wei Liu, Yangqing Jia, Pierre Sermanet, Scott Reed, Dragomir
  Anguelov, Dumitru Erhan, Vincent Vanhoucke, and Andrew Rabinovich.
\newblock Going deeper with convolutions.
\newblock In {\em Proceedings of the IEEE conference on computer vision and
  pattern recognition}, pages 1--9, 2015.

\bibitem{terhorst2020beyond}
Philipp Terh{\"o}rst, Daniel F{\"a}hrmann, Naser Damer, Florian Kirchbuchner,
  and Arjan Kuijper.
\newblock Beyond identity: What information is stored in biometric face
  templates?
\newblock In {\em 2020 ieee international joint conference on biometrics
  (ijcb)}, pages 1--10. IEEE, 2020.

\bibitem{wang2021representative}
Chengrui Wang and Weihong Deng.
\newblock Representative forgery mining for fake face detection.
\newblock In {\em Proceedings of the IEEE/CVF Conference on Computer Vision and
  Pattern Recognition}, pages 14923--14932, 2021.

\bibitem{wang2018cosface}
Hao Wang, Yitong Wang, Zheng Zhou, Xing Ji, Dihong Gong, Jingchao Zhou, Zhifeng
  Li, and Wei Liu.
\newblock Cosface: Large margin cosine loss for deep face recognition.
\newblock In {\em Proceedings of the IEEE conference on computer vision and
  pattern recognition}, pages 5265--5274, 2018.

\bibitem{wang2020mitigate}
Mei Wang and Weihong Deng.
\newblock Mitigating bias in face recognition using skewness-aware
  reinforcement learning.
\newblock In {\em 2020 IEEE/CVF Conference on Computer Vision and Pattern
  Recognition (CVPR)}, pages 9319--9328, 2020.

\bibitem{wang2021deep}
Mei Wang and Weihong Deng.
\newblock Deep face recognition: A survey.
\newblock {\em Neurocomputing}, 429:215--244, 2021.

\bibitem{wang2019racial}
Mei Wang, Weihong Deng, Jiani Hu, Xunqiang Tao, and Yaohai Huang.
\newblock Racial faces in the wild: Reducing racial bias by information
  maximization adaptation network.
\newblock In {\em Proceedings of the ieee/cvf international conference on
  computer vision}, pages 692--702, 2019.

\bibitem{wang2021meta}
Mei Wang, Yaobin Zhang, and Weihong Deng.
\newblock Meta balanced network for fair face recognition.
\newblock {\em IEEE transactions on pattern analysis and machine intelligence},
  2021.

\bibitem{wu2018towards}
Zhenyu Wu, Zhangyang Wang, Zhaowen Wang, and Hailin Jin.
\newblock Towards privacy-preserving visual recognition via adversarial
  training: A pilot study.
\newblock In {\em Proceedings of the European Conference on Computer Vision
  (ECCV)}, pages 606--624, 2018.

\bibitem{xu2021consistent}
Xingkun Xu, Yuge Huang, Pengcheng Shen, Shaoxin Li, Jilin Li, Feiyue Huang,
  Yong Li, and Zhen Cui.
\newblock Consistent instance false positive improves fairness in face
  recognition.
\newblock In {\em Proceedings of the IEEE/CVF Conference on Computer Vision and
  Pattern Recognition}, pages 578--586, 2021.

\bibitem{yi2014learning}
Dong Yi, Zhen Lei, Shengcai Liao, and Stan~Z Li.
\newblock Learning face representation from scratch.
\newblock {\em arXiv preprint arXiv:1411.7923}, 2014.

\bibitem{yin2019feature}
Xi Yin, Xiang Yu, Kihyuk Sohn, Xiaoming Liu, and Manmohan Chandraker.
\newblock Feature transfer learning for face recognition with under-represented
  data.
\newblock In {\em Proceedings of the IEEE/CVF conference on computer vision and
  pattern recognition}, pages 5704--5713, 2019.

\bibitem{zhang2017range}
Xiao Zhang, Zhiyuan Fang, Yandong Wen, Zhifeng Li, and Yu Qiao.
\newblock Range loss for deep face recognition with long-tailed training data.
\newblock In {\em Proceedings of the IEEE International Conference on Computer
  Vision}, pages 5409--5418, 2017.

\end{thebibliography}
}

\end{document}